\documentclass[review]{elsarticle}

\usepackage{hyperref}
\usepackage{amsmath} % assumes amsmath package installed
\usepackage{amssymb}  % assumes amsmath package installed
\usepackage{mathtools}
\usepackage{ctable} % for \specialrule command
\usepackage{graphicx} % for pdf, bitmapped graphics files
\DeclarePairedDelimiter\ceil{\lceil}{\rceil}
\usepackage{multirow}
\usepackage{multicol}
\usepackage{ctable} % for \specialrule command

\usepackage{subfigure}
\graphicspath{{./pics/}}

\journal{https://doi.org/10.1016/j.eswa.2021.114906}

\biboptions{authoryear}

\begin{document}

\begin{frontmatter}

\title{WiFiNet: WiFi-based indoor localisation using CNNs}%{Elsevier \LaTeX\ template\tnoteref{mytitlenote}}
% \tnotetext[mytitlenote]{Fully documented templates are available in the elsarticle package on \href{http://www.ctan.org/tex-archive/macros/latex/contrib/elsarticle}{CTAN}.}

\author[uah]{Noelia Hern\'andez\corref{cor1}}
\cortext[cor1]{Corresponding author}
\ead{noelia.hernandez@uah.es}

\author[uah]{Ignacio Parra}
\ead{ignacio.parra@uah.es}

\author[uah]{H\'ector Corrales}
\ead{hector.corrales@uah.es}

\author[uah]{Rub\'en Izquierdo}
\ead{ruben.izquierdo@uah.es}

\author[uah]{Augusto Luis Ballardini}
\ead{augusto.ballardini@uah.es}

\author[uah]{Carlota Salinas}
\ead{carlota.salinasmaldo@uah.es}

\author[uah]{Iv\'an Garc\'ia}
\ead{ivan.garciad@uah.es}

\address[uah]{Universidad de Alcal\'a, 28805 Alcal\'a de Henares (Madrid), Spain}

% \author[mymainaddress,mysecondaryaddress]{Elsevier Inc}
% \author[mysecondaryaddress]{Global Customer Service\corref{mycorrespondingauthor}}
% \cortext[mycorrespondingauthor]{Corresponding author}
% \address[mymainaddress]{1600 John F Kennedy Boulevard, Philadelphia}
% \address[mysecondaryaddress]{360 Park Avenue South, New York}

\begin{abstract}
% Graphical abstract
% Although a graphical abstract is optional, its use is encouraged as it draws more attention to the online
% article. The graphical abstract should summarize the contents of the article in a concise, pictorial form
% designed to capture the attention of a wide readership. Graphical abstracts should be submitted as a
% separate file in the online submission system. Image size: Please provide an image with a minimum
% of 531 × 1328 pixels (h × w) or proportionally more. The image should be readable at a size of 5 ×
% 13 cm using a regular screen resolution of 96 dpi. Preferred file types: TIFF, EPS, PDF or MS Office
% files. You can view Example Graphical Abstracts on our information site.
% Authors can make use of Elsevier's Illustration Services to ensure the best presentation of their images
% and in accordance with all technical requirements.
% 
Different technologies have been proposed to provide indoor localisation: magnetic field, bluetooth , WiFi, etc. Among them, WiFi is the one with the highest availability and highest accuracy. This fact allows for an ubiquitous accurate localisation available for almost any environment and any device. However, WiFi-based localisation is still an open problem.

In this article, we propose a new WiFi-based indoor localisation system that takes advantage of the great ability of Convolutional Neural Networks in classification problems. Three different approaches were used to achieve this goal: a custom architecture called WiFiNet designed and trained specifically to solve this problem and the most popular pre-trained networks using both transfer learning and feature extraction. 

Results indicate that WiFiNet is as a great approach for indoor localisation in a medium-sized environment (30 positions and 113 access points) as it reduces the mean localisation error (33\%) and the processing time when compared with state-of-the-art WiFi indoor localisation algorithms such as SVM. 

\end{abstract}


\end{frontmatter}

\newpage
%\linenumbers

\section{Introduction}

\begin{figure}[!htp]
  \centering
  \includegraphics[width=0.85\linewidth]{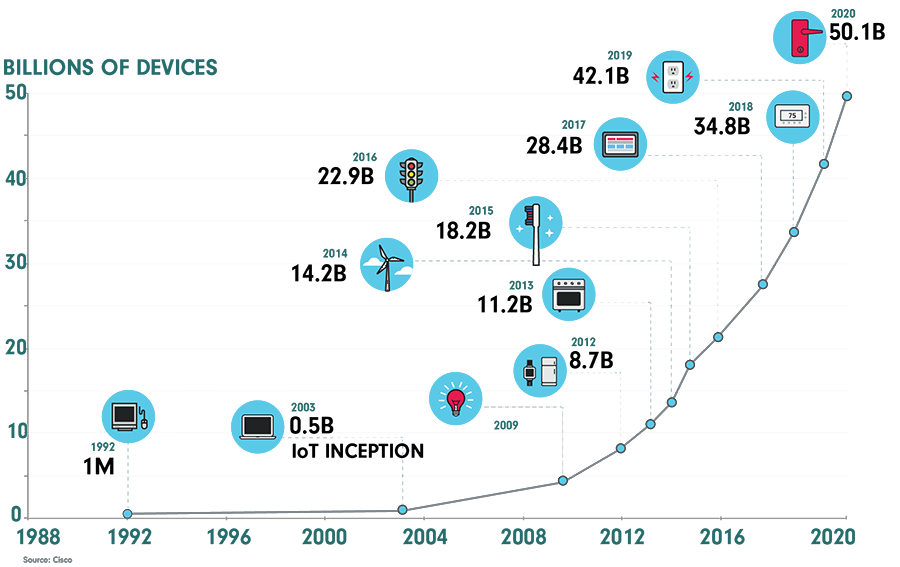}
  \caption{Growth of IoT devices. Source: Cisco.}
  \label{fig:IoT}
\end{figure}

In the recent years, the number of applications and connected devices have grown exponentially (Figure \ref{fig:IoT}). According to CISCO, this trend is expected to continue in the following years reaching 28.5 billion IoT (Internet of Things) connected devices in 2022 \citep{references:cisco}. Laptops, phones and tablets were the devices traditionally used for connectivity. However, in the recent years lots of other devices, such as smartwatches, fridges, even bulbs or toothbrushes, have joined the world of connected devices. This trend seeks to help the users in every aspect of their life \citep{references:iot} with IoT applications in smart homes \citep{references:smarthomes}, cities \citep{references:smartcities} or warehouses \citep{references:warehouses}.

With smart sensors almost everywhere, IoT applications will require or benefit from a reliable and accurate localisation of certain devices. For instance, locating medical staff, medical equipment or even patients in hospitals would increase the quality and rapidity of the medical services \citep{references:hospitals}. This localisation must be real-time, accurate both indoors and outdoors \citep{RACKO201888} and with low power consumption. So far, this localisation was generally provided through the use of GPS (Global Positioning System) which has one main problem: it does not work indoors because of the NLOS (Non-line-of-sight) effect.

As a consequence, different technologies have been proposed to provide indoor localisation \citep{references:indoorSurvey}: magnetic field \citep{references:98}, Bluetooth \citep{references:bluetooth1, references:bluetooth2}, WiFi \citep{references:WiFi1,references:WiFi2}, UWB (Ultra Wide Band) \citep{references:UWB}, RFID \citep{references:rfid} or visible light \citep{references:light1,references:light2} among others.

Wireless RSSI-based indoor localisation (based on Bluetooth, UWB or WiFi) is the most common due to its high availability (in both IoT devices and the environments), low consumption and high accuracy. Among them, WiFi is the one with the highest availability and highest accuracy at the cost of slightly higher power consumption \citep{references:RSSIsurvey}. This fact allows for an ubiquitous accurate localisation available for almost any environment and any device.

However, WiFi-based localisation is still an open problem. Since RADAR, the first method proposed in 2000 \citep{references:bahl00}, the research community has proposed different solutions to increase the localisation accuracy using methods based on state-of-the-art algorithms such as Random Forest \citep{references:jedari15} or classifier ensembles \citep{references:torres-sospedra16_2,references:eswa17} achieving promising results. 

But most of these methods, based on classic machine learning, have two main problems: their low scalability to big environments and their low generalisation ability which increases the localisation error when the number of site-surveyed positions is not high enough for the size of the environment. As a consequence, new methods were proposed to solve these problems seeking for a reduction of the site-survey effort \citep{references:sensors17} or a self-maintenance of the system \citep{references:Tao18}. Despite all the efforts dedicated to indoor WiFi-based localisation research, there is still room for improvement.

In this article, we propose a new WiFi-based indoor localisation system that takes advantage of the CNNs (Convolutional Neural Networks) great performance in classification problems, looking for a reduction of the localisation error without forgetting about its generalisation ability and its scalability. Three different approaches were explored to achieve this goal: a custom architecture, called WiFiNet, designed and trained specifically to solve this problem, transfer learning using eight well-known pre-trained networks and, finally, feature extraction from pre-trained networks to be used along with classic classifiers. 

As most CNNs are designed to classify images, the WiFi samples (the RSS (Received Signal Strength) from the WiFi APs (Access Points)) must be converted into images before training the network. The way the APs are ordered to form the image is key for the localisation procedure, as the CNNs will search for features taking small areas of the images as a whole. For this reason, the images will be created ordering the APs according to the order they appear in the collected data increasing its spatial relation. 
Once the images are created, they can be fed to the CNN for both training in an initial stage and classification during the use of the CNN for device localisation.

Three different experiments were designed to test the system: localisation at positions covered in the training dataset (the usual way of testing new localisation systems), localisation at positions non existent in the training dataset (to evaluate its generalisation ability) and localisation under realistic conditions (walking around the environment). Test data was collected on a different time and date from the training data to capture the high variability of the WiFi signal over time. In addition, the processing time is analysed to evaluate the scalability of the system.

WiFiNet arises as a great approach for indoor localisation in a medium-sized environment (30 positions and 113 APs) as it reduces the mean localisation error when compared with state-of-the-art algorithms (RMSE of 3.3\,m compared with 4.4\,m using SVM). At the same time, WiFiNet's scalability lead us to think that it will be able to perform real-time localisation in bigger environments than state-of-the-art algorithms.

The rest of the manuscript is structured as follows: Section \ref{section:method} presents the core of this work. It describes the different approaches to work with CNNs and describes WiFiNet, our custom architecture. Then, Section \ref{section:results} goes in depth with the experimental evaluation, analisys of the results and discussion. Finally, the main contributions in this work along with some future research lines are summarized in Section \ref{section:conclusions}.

\section{CNNs for WiFi indoor localisation}\label{section:method}

% Provide sufficient details to allow the work to be reproduced by an independent researcher. Methods that are already published should be summarized, and indicated by a reference. If quoting directly from a previously published method, use quotation marks and also cite the source. Any modifications to existing methods should also be described.

Recently, CNNs, a special kind of multi-layer neural network including convolutional layers, have become a key method for classification specially in computer vision problems.  In this area, CNNs have beaten traditional methods to the point that they are decreasingly being used for classification problems. 

However, CNNs goodness is infrequently tested in other areas. When training a CNN to solve a new problem, there are four different approaches that might be used:

\begin{itemize}

    \item Design and train a new network architecture: This approach requires a deep understanding of CNNs architecture design, a big dataset and powerful hardware to train the new network. This method is usually chosen when the problem to solve is very different from the problems solved by the already existing networks and the number of available training images is high enough.

    \item Use of a pre-trained network (transfer learning): There is a wide range of pre-trained networks that can be used as a starting point to solve new problems. Most of these models have in common that they were designed to improve the performance classifying Imagenet \citep{imagenet}, a large scale dataset (more than a million images) with 1000 different classes used in the ImageNet Large-Scale Visual Recognition Challenge (ILSVRC) \citep{ILSVRC}.  In this approach, the knowledge learned during the previous training (low-level features extraction such as colors, edges, shapes, etc.) is maintained and only a fine-tunning will be done during the new training. By using transfer learning, the new training will be much faster and the number of images required in the new dataset much smaller than training a network from scratch.

    \item Train an existing network architecture from scratch: This approach re-uses the design of an already existing CNN, but the weights are re-computed to solve the new problem. The difference with the first approach is that there is no need to design a CNN architecture, but a high number of images are still needed to train the network. This approach is generally used when the problem is similar to an existing one but the previous method (transfer learning) is not providing the expected results.

    \item Use an existing pre-trained network to extract features and then using them as inputs for a classic classification algorithm: This approach takes advantage of the architecture and weights of pre-trained networks. The CNN is only used to extract features that will be used as input features for other classifiers (such as KNN or SVM). This approach is mostly used when the computing power, the time or the number of samples is insufficient to train a network but also to take advantage of the great ability of CNNs extracting significant features. 

\end{itemize}

In this article, we propose to use CNNs to estimate the location of a device by using the RSS from the WiFi APs on the environment. To do so, we have followed three approaches: A new network architecture was designed and trained from scratch, different pre-trained networks were used to take advantage of transfer learning and finally, pre-trained networks were also used to extract features to be used as inputs with some classic classifiers. 

As all of these networks are designed to classify images, the first step will be to transform the RSS samples. Once the RSS samples are transformed into images, they can be used to train the network (or extract features). Naturally, the same transformation will be required to locate the device during the localisation stage. 

\subsection{Creation of images from RSS samples}\label{sec:imagesCreation}

The first step to create the images from the RSS samples is to select the order in which the APs will be arranged in the images. This is not a trivial procedure, as the CNNs will apply the operations through the different layers on spatially-neighboring pixels, especially in their initial layers \citep{references:coates12}. As a consequence, RSS information from the different APs will change from being unrelated to being associated with their neighbouring RSS. In this paper, the order of the APs is selected according to the order they appear in the collected data increasing its spatial relation. This way, the APs that are visible from the same position of the environment will be arranged on the same area of the image creating groups of neighbouring pixels with related information. Then, the data of each sample will be re-arranged following the selected order into an Y by Y square matrix, with
\begin{equation}
    Y = \ceil*{\sqrt{N_{AP}}}
\end{equation}
being $N_{AP}$ the number of APs in the complete training dataset. 

The next step is to transform the RSS values into pixel intensity values (i.e. from 0 to 255). Given that the collected RSS ranges from -99\,dBm to -30\,dBm, this transformation is performed by adding an offset of +200 to the RSS, obtaining medium-ranged intensity values (from 101 to 170). If there is no RSS from an AP, its pixel intensity will be set to 0. This way, there is a high differentiation between non-seen APs and low RSS APs. Of course, if the total number of APs is not a perfect square, the image will have some pixels without an associated AP which will also be set to  0.

\begin{figure}[!htp]
  \centering
  \includegraphics[width=\linewidth]{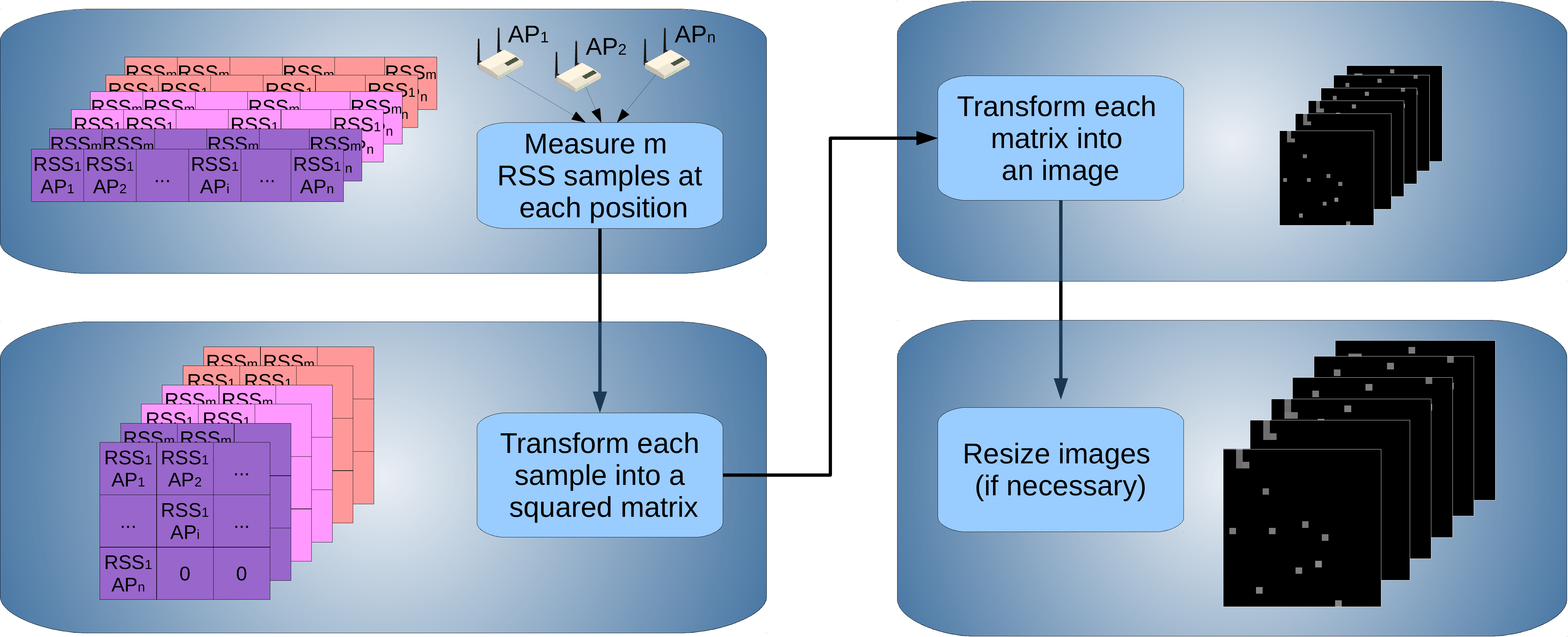}
  \caption{Collecting and pre-processing data to convert RSS samples into images.}
  \label{fig:imagesGeneration}
\end{figure}

Finally, depending on the number of APs and the selected CNN, the images should be resized to agree with the input layer of the network (pre-trained networks input image size is around 224x224 pixels). Once the training RSS samples are transformed into images, they can be used to train a network following one of the approaches explained at the beginning of section \ref{section:method}. The complete process is summarized in Figure \ref{fig:imagesGeneration}.

\subsection{WiFiNet: a custom architecture}

As explained at the beginning of the section, custom architectures are used when the problem to solve is highly different from the problems already solved by the existing networks and the number of available training samples is enough. However, knowing how many samples are enough to train a network in advance is almost impossible, being necessary to train and test the network to obtain this knowledge from the results. Thus, in practice, this method is used to solve new problems as is our case: Existing architectures and pre-trained networks are designed with the aim of identifying real objects (cars, animals, etc.) in the input images, but our goal is to obtain a location from RSS measurements which have no evident meaning in the image.

\begin{figure}[!htp]
  \centering
  \includegraphics[width=\linewidth]{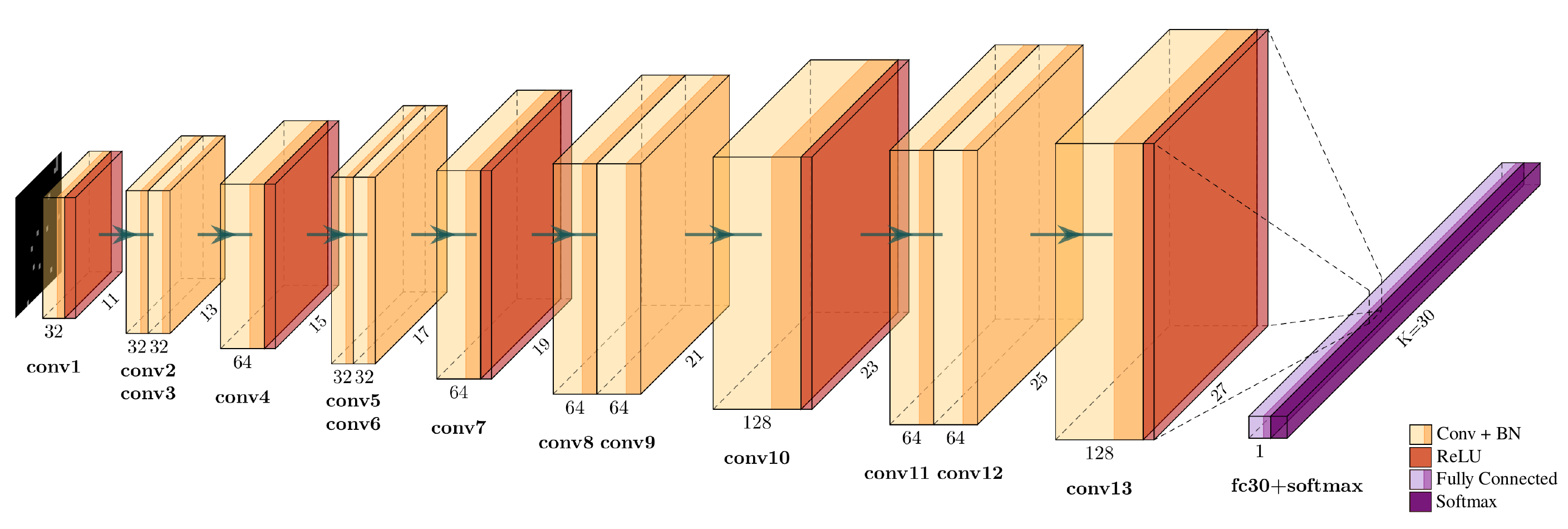}
  \caption{WiFiNet: Custom architecture.}
  \label{fig:WiFiNet}
\end{figure}

In this work, we propose a custom network architecture, called WiFiNet, based on ResNet bottleneck units to take advantage of its efficiency and accuracy. Our network is formed by 13 convolutional layers (Conv) followed by a Batch Normalisation (BN) layer and a ReLU layer (Rectified Linear Unit) every three Conv+BN layers. Figure\,\ref{fig:WiFiNet} shows the complete architecture. As it can be seen, an initial input Conv+BN+ReLU block is followed by four Conv+BN+Conv+BN+Conv+BN+ReLU blocks in which the output of the different layers grows with their depth in the network. This decision was made to keep the weight of the pixels on the borders of the images the same as the rest of the pixels. This decision was validated experimentally as this network got the best results in terms of accuracy, mean error and processing time. The last layer of the network is a Fully Connected+Softmax layer that will use the activations of the previous layer to decide the most likely position in which the RSS sample was collected. As this network was specifically designed to solve this localisation problem, there is no need to resize the input images (as explained section\,\ref{sec:imagesCreation}). 

\subsection{Transfer learning using pre-trained networks}

Transfer learning (TL) is a great approach to reduce the training effort (in time, processing power and training dataset size) by using a pre-trained network as starting point to learn how to solve new problems. In this approach, the initial layers of the network are maintained along with their weights to keep the knowledge obtained during the previous training (low-level features extraction such as colors, edges, shapes, etc.). The weights of these layers will be fine-tuned during the new training. The final layers of the network (classification layers) are replaced with ``blank'' ones configured to differentiate between the classes of the new problem. The weights of the classification layers will be adjusted from scratch during the training.

There are lots of pre-trained CNNs for classification that can be used for this purpose. The most popular ones are AlexNet \citep{alexnet}, GoogLeNet \citep{googlenet}, ResNet \citep{resnet} or SqueezeNet \citep{SqueezeNet} but others such as NasNet Mobile \citep{NasNetMobile} or Inception ResNetv2 \citep{inceptionResNet} are deeper networks that are also worth testing. Their input layer is designed to use 224x224 or 227x227 square images. As a consequence, the input images must be resized to fulfill this requirement. 

A summary of all the TL algorithms and configurations is shown in table \ref{tab:algorithms} where sgdm is the Stochastic Gradient Descent with Momentum optimizer, LR is the Learning Rate and MB is the size of the mini-batch for each training iteration. As can be seen, the selected parameters are the same for all the CNNs, except for ResNet101, NasNet Mobile and Inception ResNetv2 in which the selected mini-batch size is smaller in order to fit the network in the GPU memory. Changing this parameter may lead to better generalization at the cost of higher computational time per epoch during the training.

\begin{table}[!ht]
%   \renewcommand{\arraystretch}{1.5}
  \caption{Summary of the algorithms and their configuration.}
  \label{tab:algorithms}
  \begin{center}
    \begin{tabular}{ccc} 
      \specialrule{.2em}{.1em}{.1em} 
	  Algorithm & Short name & Configuration \\
       \specialrule{.2em}{.1em}{.1em} 
	 WiFiNet  & WiFiNet & sgdm, $10^{-3}$ LR, 120 MB \\
       
	 AlexNet (TL) \citep{alexnet}  & AlexNet & sgdm, $10^{-3}$ LR, 120 MB \\
       
	 SqueezeNet (TL) \citep{SqueezeNet}  & Squeeze & sgdm, $10^{-3}$ LR, 120 MB \\
       
	 ResNet18 (TL) \citep{resnet}  & ResN18 & sgdm, $10^{-3}$ LR, 120 MB \\
       
	 ResNet50 (TL) \citep{resnet}  & ResN50 & sgdm, $10^{-3}$ LR, 120 MB \\
       
	 ResNet101 (TL) \citep{resnet}  & ResN101 & sgdm, $10^{-3}$ LR, 60 MB \\
       
	 GoogLeNet (TL) \citep{googlenet}  & GoogLe & sgdm, $10^{-3}$ LR, 120 MB \\
       
	 NasNet Mobile (TL) \citep{NasNetMobile} & NasMob & sgdm, $10^{-3}$ LR, 60 MB \\
       
	 Inception ResNetv2 (TL) \citep{inceptionResNet} & Inception & sgdm, $10^{-3}$ LR, 30 MB \\
      
	 SVM (AlexNet FE) & SVMAlex & Linear kernel  \\
      
	 SubKNN (AlexNet FE) & sKNNAlex & Euclidean distance, K=1  \\
      
	 SVM (ResNet18 FE) & SVMRN18 & Linear kernel  \\
      
	 SubKNN (ResNet18 FE) & sKNNRN18 & Euclidean distance, K=1   \\
       
	 KNN \citep{references:KNN} & KNN & Euclidean distance, K=1  \\
       
	 SVM \citep{references:SVM} & SVM & Linear kernel  \\
      
	 Bagged Trees \citep{references:bagged} & BagT & MSE split criterion  \\
      
	 AdaBoost \citep{references:adaboost} & BoostT & MSE split criterion  \\
      
	 Subspace Discriminant \citep{references:subspace} & SubDisc & Linear Discriminant  \\
      
	 Subspace KNN \citep{references:subspace} & SubKNN & Euclidean distance, K=1  \\
      \specialrule{.2em}{.1em}{.1em} 
    \end{tabular}
  \end{center}
\end{table}

\subsection{Feature extraction from pre-trained networks}

Feature extraction (FE) also takes advantage of the knowledge learned by a pre-trained network, but reduces the time, processing power and dataset size to a minimum since there is no need to retrain the network. With this approach, the activations (features) of the layer previous to the classification layers are directly used to train a classic classifier.

The combinations of two pre-trained CNNs with two classic classifiers were used to test this approach: ResNet18 and AlexNet were used to extract the features and then, SVM and subspaceKNN were trained using the extracted features. Table \ref{tab:algorithms} also contains the algorithms and configurations using feature extraction.

\subsection{Classic classifiers: baseline}

Finally, six machine learning algorithms were tested for comparison purposes. Two classic classifiers: K-Nearest Neighbours (KNN) \citep{references:KNN} and Support Vector Machines (SVM) \citep{references:SVM} and four Classifier Ensembles (CE): AdaBoost \citep{references:adaboost}, Bagged Trees \citep{references:bagged}, Subspace KNN and Subspace Discriminant \citep{references:subspace}. The configuration of these algorithms can be found on table \ref{tab:algorithms}.

\section{Results and Discussion}\label{section:results}

This section presents the experiments carried out to validate the localisation algorithms. First, the experimental set-up is described. Then, the different experiments along with a critical discussion of the obtained results is provided.

Three different experiments are presented: training and test using measurements collected at the same locations (test data collected one week apart from training data), test using measurements collected at different positions from the training ones (collected two months apart from training data) and test using data collected in motion (collected two months apart from training data).

\subsection{Experimental set-up}

\begin{figure}[!hpt]
  \centering
  \mbox{		
		\subfigure[Training positions]{
			\includegraphics[width=0.47\linewidth] {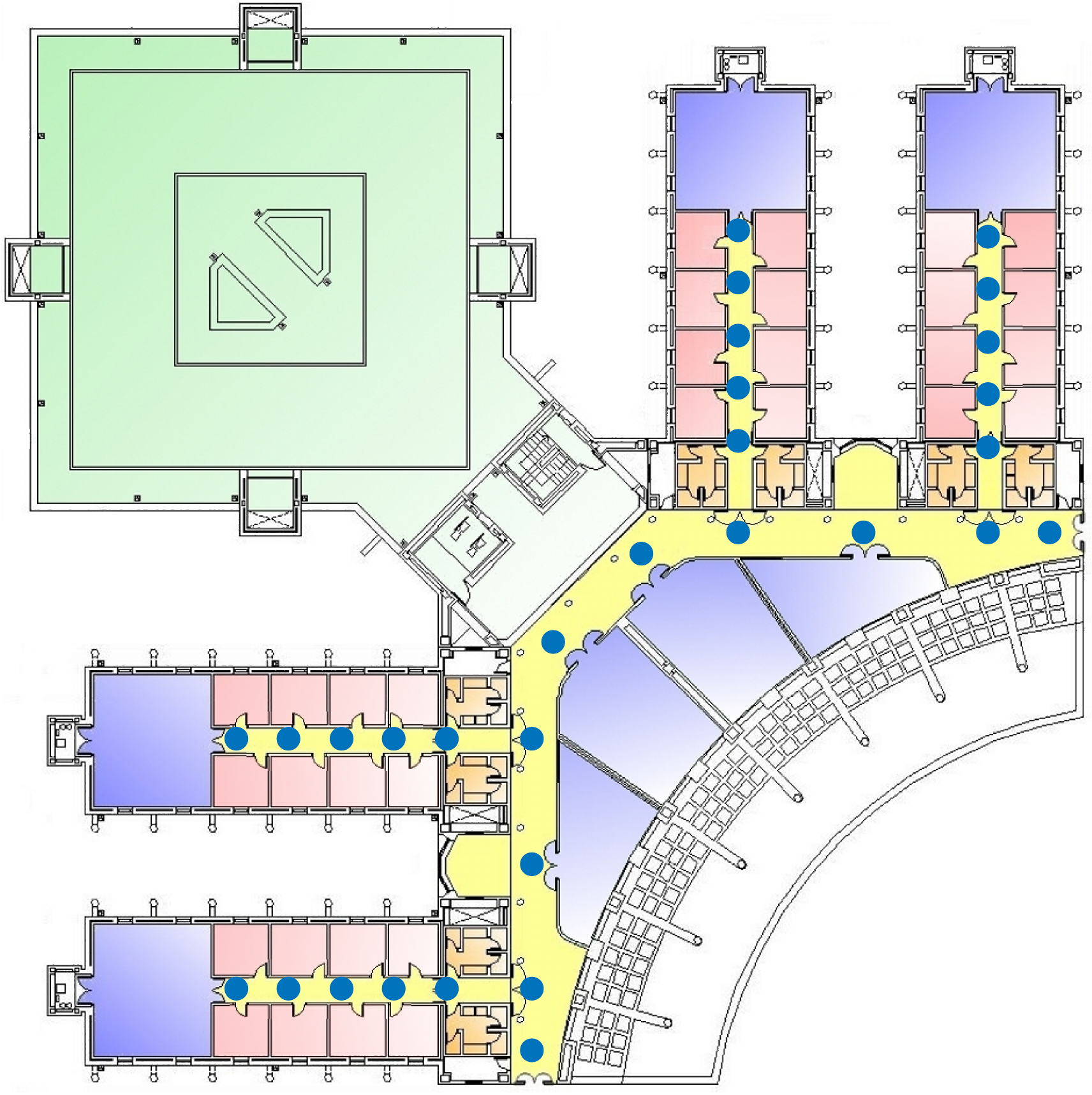}
			\label{subfig:trainPos}
		}	
		\subfigure[Test positions]{
			\includegraphics[width=0.47\linewidth] {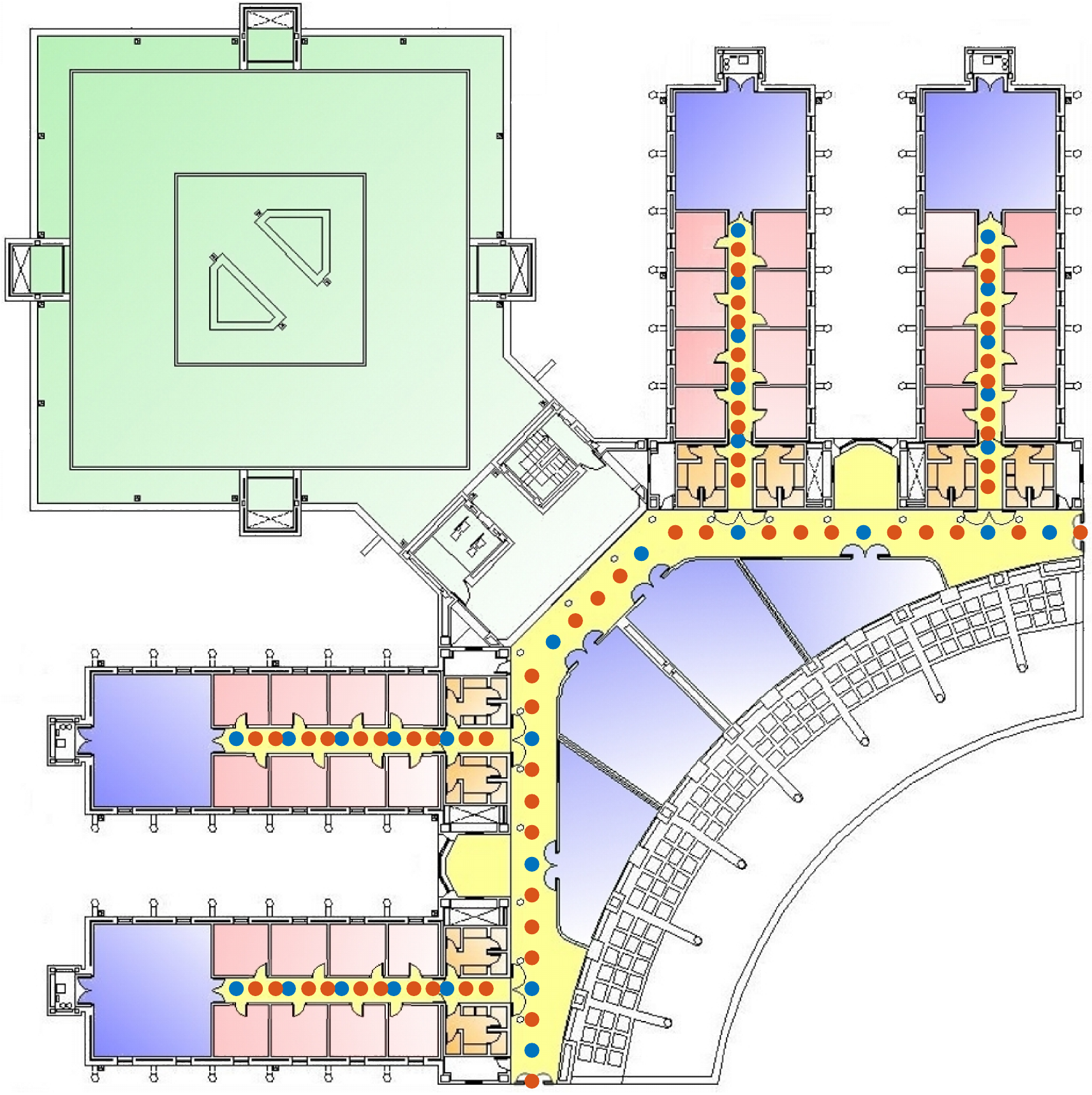}
			\label{subfig:testPos}
		}			
  }
%   \mbox{		
		\subfigure[Trajectory]{
			\includegraphics[width=0.47\linewidth] {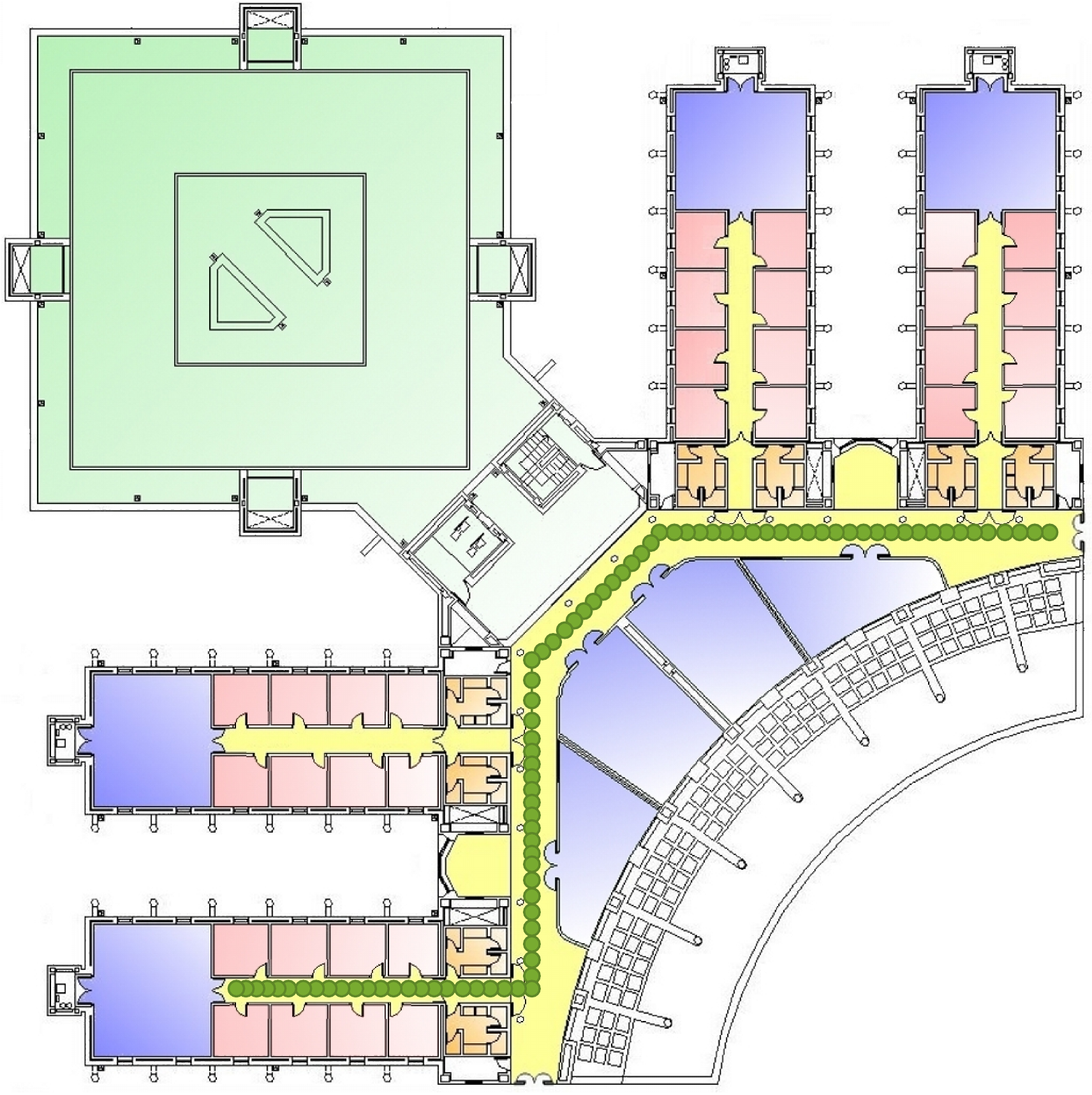}
			\label{subfig:traj}
		}			
%   }
  \caption{Experimental environment located at the University of Alcal\'a. Training positions are represented with blue circles. Test positions different from the training ones are represented using red circles.}
  \label{fig:Environment}
\end{figure}

The proposed CNNs were tested in a complex real-world environment set at the University of Alcal\'a (Madrid, Spain) covering 3600\,$m^2$ (Figure\,\ref{fig:Environment}). 

The training dataset was collected using the internal WiFi interface of a Toshiba Port\'eg\'e Z830-10F laptop on 30 evenly distributed positions (Figure\,\ref{subfig:trainPos}). The minimum distance between neighbour positions in the training dataset is 3 m, the maximum distance 7.77 m and the mean distance 4.46 m. 

Test data was collected a week later on the training positions. The test dataset is completed with measurements collected two months after the training dataset at 67 new positions (Figure\,\ref{subfig:testPos}) and with data collected while walking around the environment (Figure\,\ref{subfig:traj}). This temporal difference on the collected data allows to capture the high variability of the WiFi signal over time. At the same time, measurements collected in non existent positions in the training dataset and during the trajectory allows to test the generalisation capability of the proposed localisation method under more realistic conditions both standing at fixed positions and in motion.

\begin{figure}[!hpt]
  \centering
  \mbox{		
		\subfigure[Pos 1 - Train]{
			\includegraphics[width=0.25\linewidth] {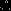}
			\label{subfig:trainPos1}
		}	
		\subfigure[Pos 5 - Train]{
			\includegraphics[width=0.25\linewidth] {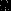}
			\label{subfig:trainPos15}
		}		
		\subfigure[Pos 13 - Train]{
			\includegraphics[width=0.25\linewidth] {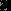}
			\label{subfig:trainPos30}
		}
  }
  \mbox{	

		\subfigure[Pos 1 - Test]{
			\includegraphics[width=0.25\linewidth] {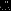}
			\label{subfig:testPos1}
		}

		\subfigure[Pos 5 - Test]{
			\includegraphics[width=0.25\linewidth] {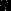}
			\label{subfig:testPos15}
		}
		
		\subfigure[Pos 13 - Test]{
			\includegraphics[width=0.25\linewidth] {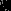}
			\label{subfig:testPos30}
		}	
  }
  \caption{Example of training and test images.}
  \label{fig:CNNimages}
\end{figure}

A total of 113 APs were detected during the collection of the training dataset. We have no information about their configuration nor their location. This way, the information from all the existing APs is used to create the input images resulting in $11\times11$ square images. Figure \ref{fig:CNNimages} shows some examples of the images used to train and test the CNN. 

\subsection{Localization at existing positions in the training dataset}

In this experiment, the test positions are the same as the training positions. This way, the behaviour of the different algorithms is tested when the localisation is performed at known positions, which is the best-case scenario. The perfect localisation algorithm should obtain a RMSE (Root Mean Square Error) of 0\,m with an accuracy of 100\%.

\begin{figure}[!hpt]
  \centering
  \mbox{		
		\subfigure[Accuracy]{
			\includegraphics[width=0.47\linewidth] {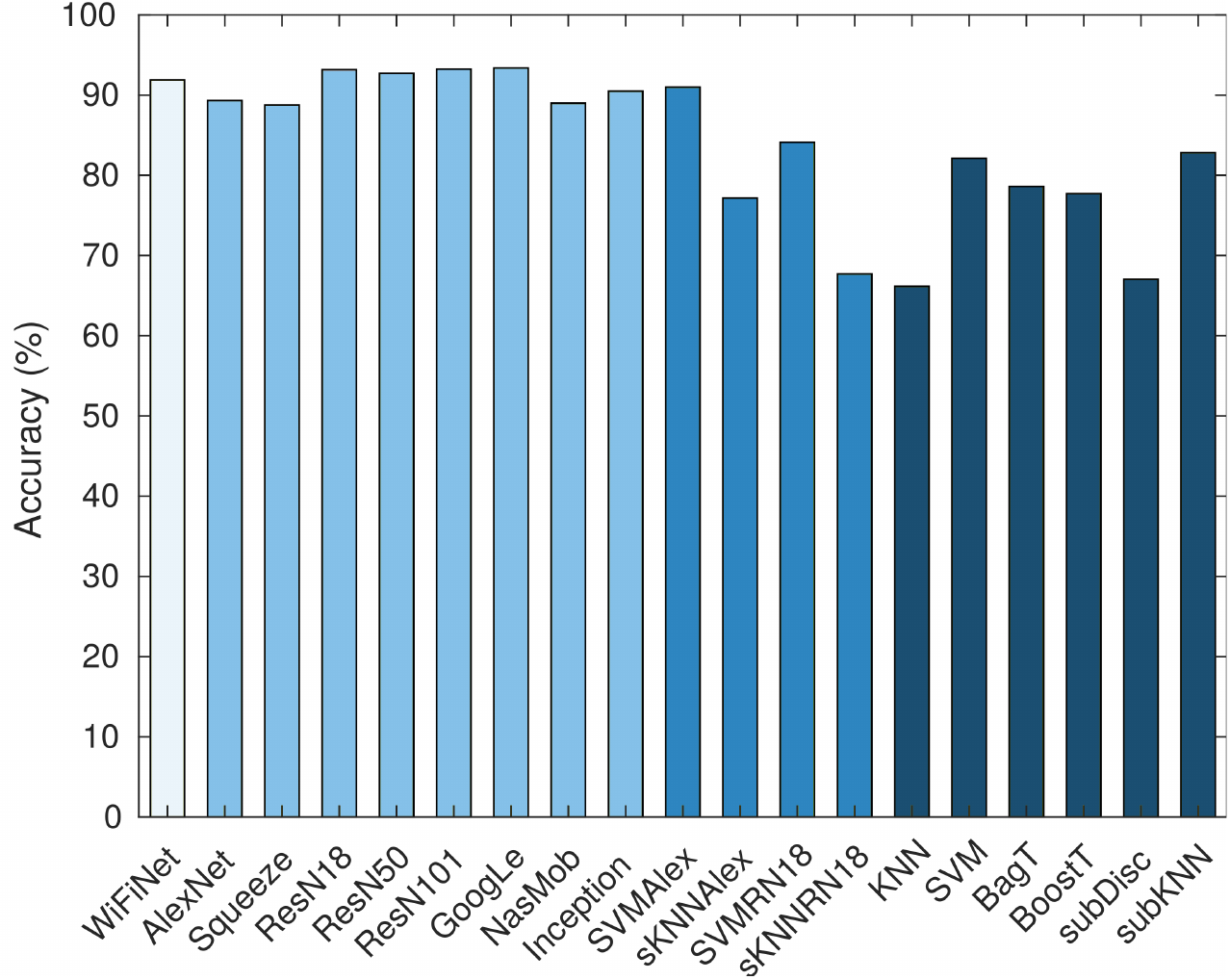}
			\label{subfig:knownAccuracy}
		}	
		\subfigure[RMSE]{
			\includegraphics[width=0.47\linewidth] {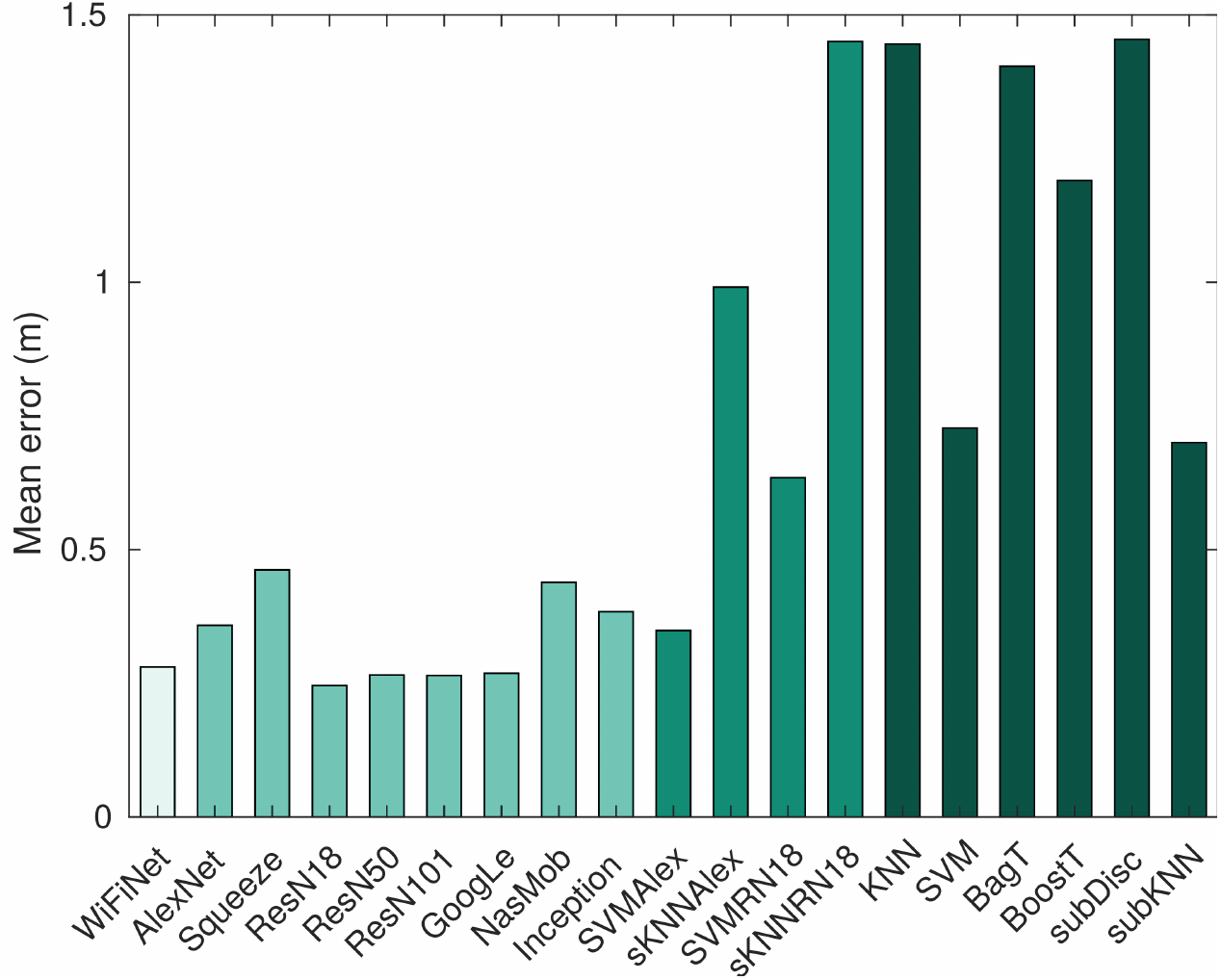}
			\label{subfig:knownMeanError}
		}	
  }
  \caption{Accuracy and RMSE using measurements collected at the same positions as the training data.}
  \label{fig:knownResults}
\end{figure}

Figure \ref{fig:knownResults} shows the accuracy (Figure \ref{subfig:knownAccuracy}) and RMSE (Figure \ref{subfig:knownMeanError}) using the algorithms summarized in table \ref{tab:algorithms}. Four different colours are used to help with the differentiation of the four different approaches (custom architecture, transfer learning, feature extraction and classic learners). Table \ref{tab:knownResults} shows a summary with the results of the best algorithms using each approach.

As can be seen, the highest accuracy and lowest RMSE is obtained using transfer learning using ResNet18 (24.6cm), closely followed by  ResNet101, ResNet50, GoogLeNet and WiFiNet (28cm). However, feature extraction using ResNet18 (63.5cm) does not seem to be as good as the use of the CNN itself. Among feature extraction algorithms, the lowest RMSE is obtained using SVM with AlexNet FE (34.9cm) while using SVM (no FE) the RMSE is almost double (72.7cm).

Looking at the results it can be concluded that WiFiNet and TL are the best approaches identifying already known locations with RMSE from 24.6cm (ResNet18) to 28cm (WiFiNet). The RMSE obtained using the best classic classifier is 250\% higher than using WiFiNet (subspaceKNN (70cm) and SVM (72.7cm)).

\begin{table}[!ht]
%   \renewcommand{\arraystretch}{1.5}
  \caption{Summary of the results locating at existing positions in the training dataset.}
  \label{tab:knownResults}
  \begin{center}
    \begin{tabular}{ccc} 
      \specialrule{.2em}{.1em}{.1em} 
	  Algorithm & Accuracy & RMSE \\
       \specialrule{.2em}{.1em}{.1em} 
	 WiFiNet  & 91.89\% & 28 cm \\
       
	 ResNet18 (TL)  & 93.17\% & 24.6 cm \\
       
	 SVMAlex (FE)  & 91\% & 35.9 cm \\
       
	 SVM  & 82.11\% & 72.7 cm\\
      \specialrule{.2em}{.1em}{.1em} 
    \end{tabular}
  \end{center}
\end{table}

\subsection{Localisation at unknown positions: Generalization ability}

This experiment is intended to evaluate the generalization ability of the different algorithms, as the test positions are different from the training ones. In this case, the accuracy of all the algorithms is always 0\,\% as the test positions do not exist in the training data and, as a consequence, the RMSE can not be 0\,m. The best performing approach will be the one with the lower RMSE.
 
\begin{figure}[!hpt]
  \centering
  \mbox{		
		\subfigure[Box plot]{
			\includegraphics[width=0.47\linewidth] {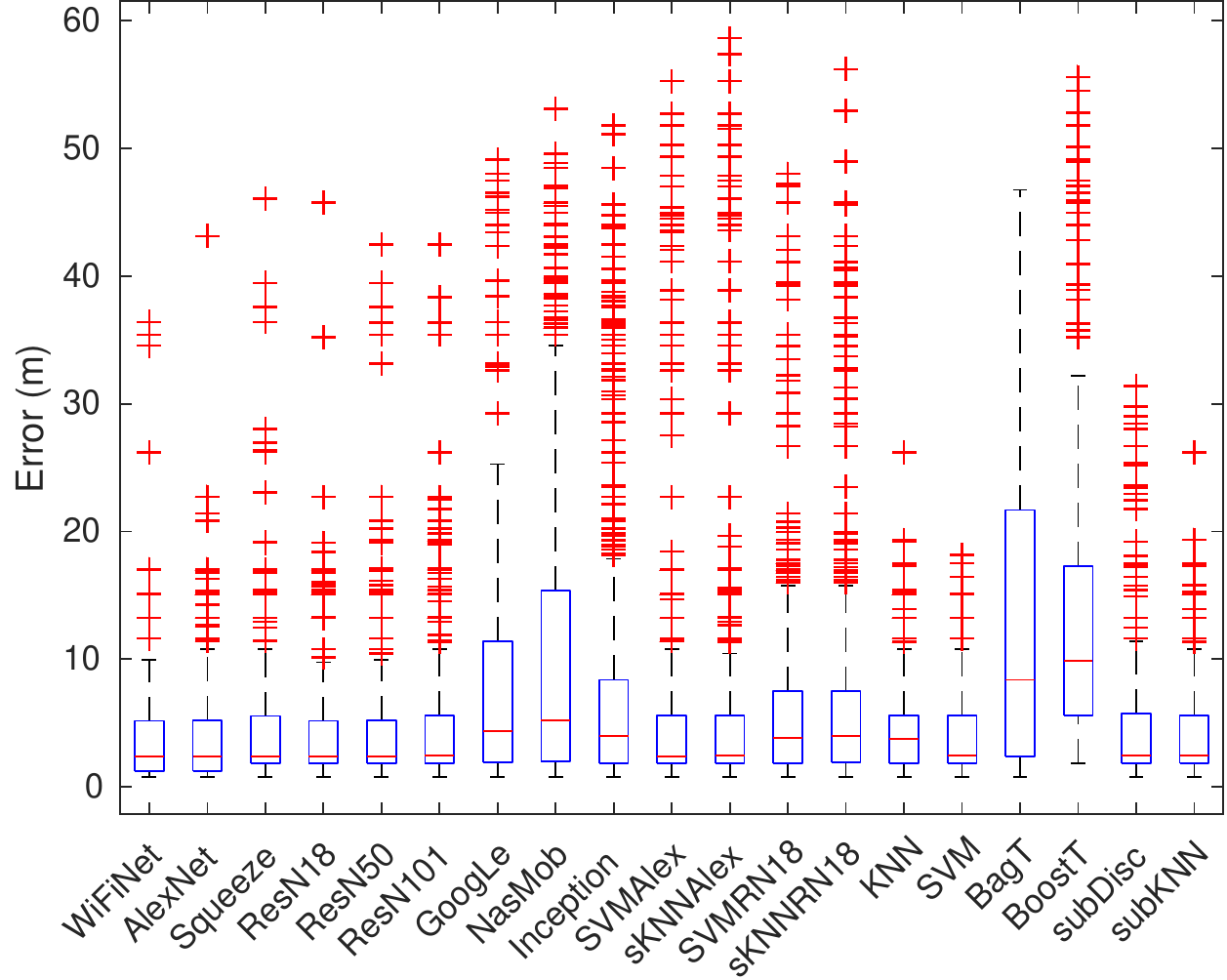}
			\label{subfig:unknownBoxplot}
		}	
		\subfigure[RMSE]{
			\includegraphics[width=0.47\linewidth] {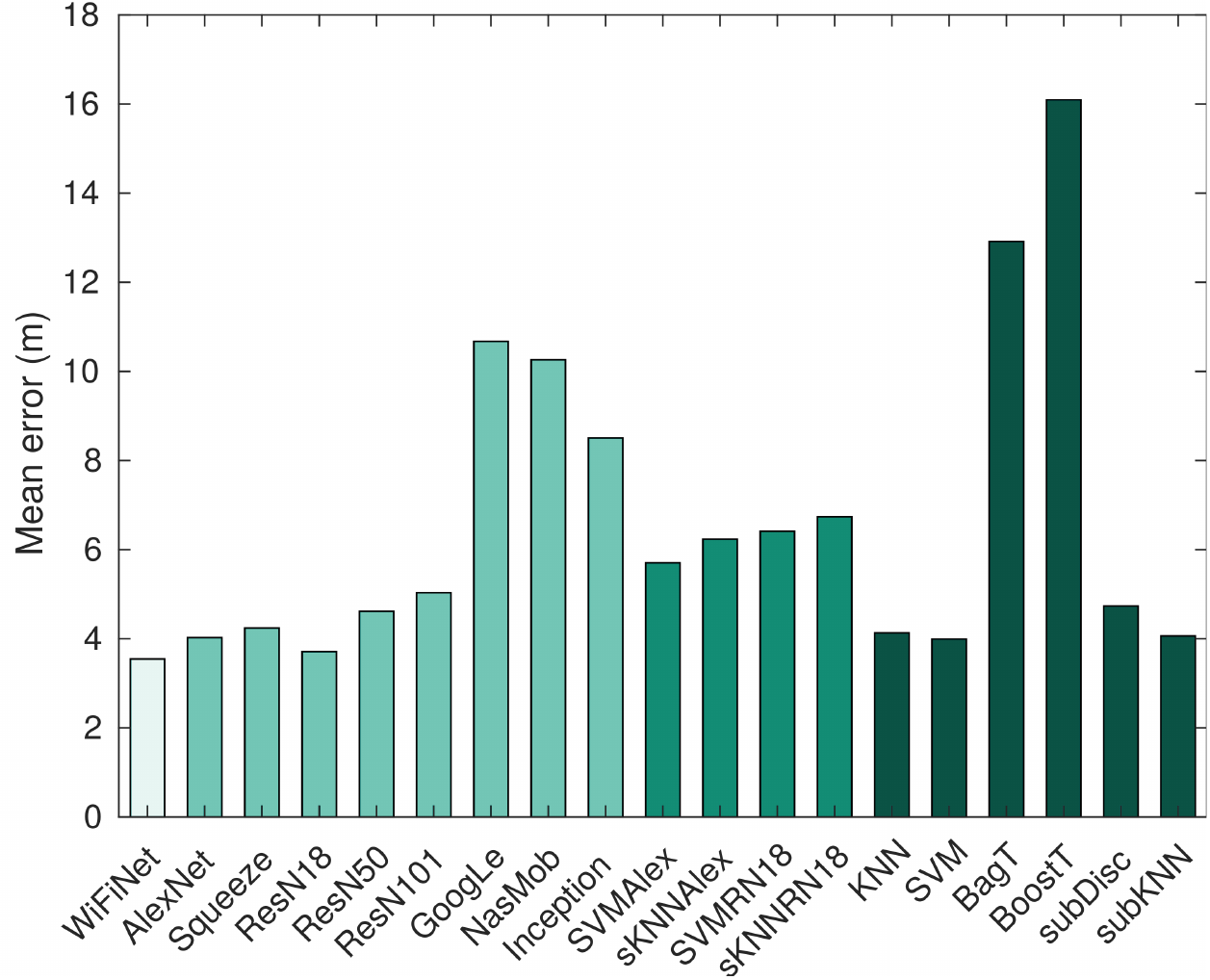}
			\label{subfig:unknownMeanError}
		}	
  }
  \caption{Box plot and RMSE using measurements collected at positions not covered by the training data.}
  \label{fig:unknownResults}
\end{figure}

Figure \ref{fig:unknownResults} shows the box plot (Figure \ref{subfig:unknownBoxplot}) and RMSE (Figure \ref{subfig:unknownMeanError}) using the algorithms summarized in table \ref{tab:algorithms}. As in the previous section, four different colours are used to help with the differentiation of the four different approaches. Table \ref{tab:unknownResults} shows a summary of the best results using each approach.

As can it be seen, the lowest RMSE and most compact box plot is now obtained using WiFiNet (3.5\,m) followed by ResNet18 (3.7\,m). The RMSE for the rest of CNNs used for TL is higher, especially in deeper networks (GoogleNet, NasNet Mobile and Inception ResNetv2) probably due to overlearning. In addition, feature extraction does not seem to be a good approach for localisation at unknown positions obtaining errors from 5.7\,m (SVM with AlexNet FE) to 6.7\,m (subspaceKNN with ResNet18 FE). Finally, SVM is the best classic learner with a RMSE of 4\,m (around a 14\% higher than using WiFiNet).

As in the previous section, WiFiNet and ResNet18 are the best locating the device at positions non existent in the training dataset. However, even though SVM with AlexNet FE was one of the best when performing the localisation at known positions, the RMSE in this experiment shows that it is not good at generalising. On the other hand, the classic classifiers performance in this experiment shows that the difference with WiFiNet and ResNet18 shortens when the localisation is performed at unknown positions even though the mean localisation error at known positions was 250\% higher than using WiFiNet.

\begin{table}[!ht]
%   \renewcommand{\arraystretch}{1.5}
  \caption{Summary of the results locating at non existent positions in the training dataset.}
  \label{tab:unknownResults}
  \begin{center}
    \begin{tabular}{ccc} 
      \specialrule{.2em}{.1em}{.1em} 
	  Algorithm & Mean Error & $75^{th}$ percentile\\
       \specialrule{.2em}{.1em}{.1em} 
	 WiFiNet  & 3.5 m & 5.1 m \\
       
	 ResNet18 (TL)  & 3.7 m & 5.2 m \\
       
	 SVMAlex (FE)  & 5.7 m & 5.6 m \\
       
	 SVM  & 4 m & 5.6 m\\
	 
      \specialrule{.2em}{.1em}{.1em} 
    \end{tabular}
  \end{center}
\end{table}

\subsection{Localisation in motion: Performance analysis of the proposed system}

Finally, this experiment is aimed to evaluate the system under realistic conditions (i.e. walking around while collecting test measurements in both known and unknown positions). To do so, the data collected following the trajectory shown in Figure \ref{subfig:traj} was used. As in the previous section, the RMSE can not be 0\,m since there are measurements collected at positions not covered in the training dataset.

\begin{figure}[!hpt]
  \centering
  \mbox{		
		\subfigure[Box plot]{
			\includegraphics[width=0.47\linewidth] {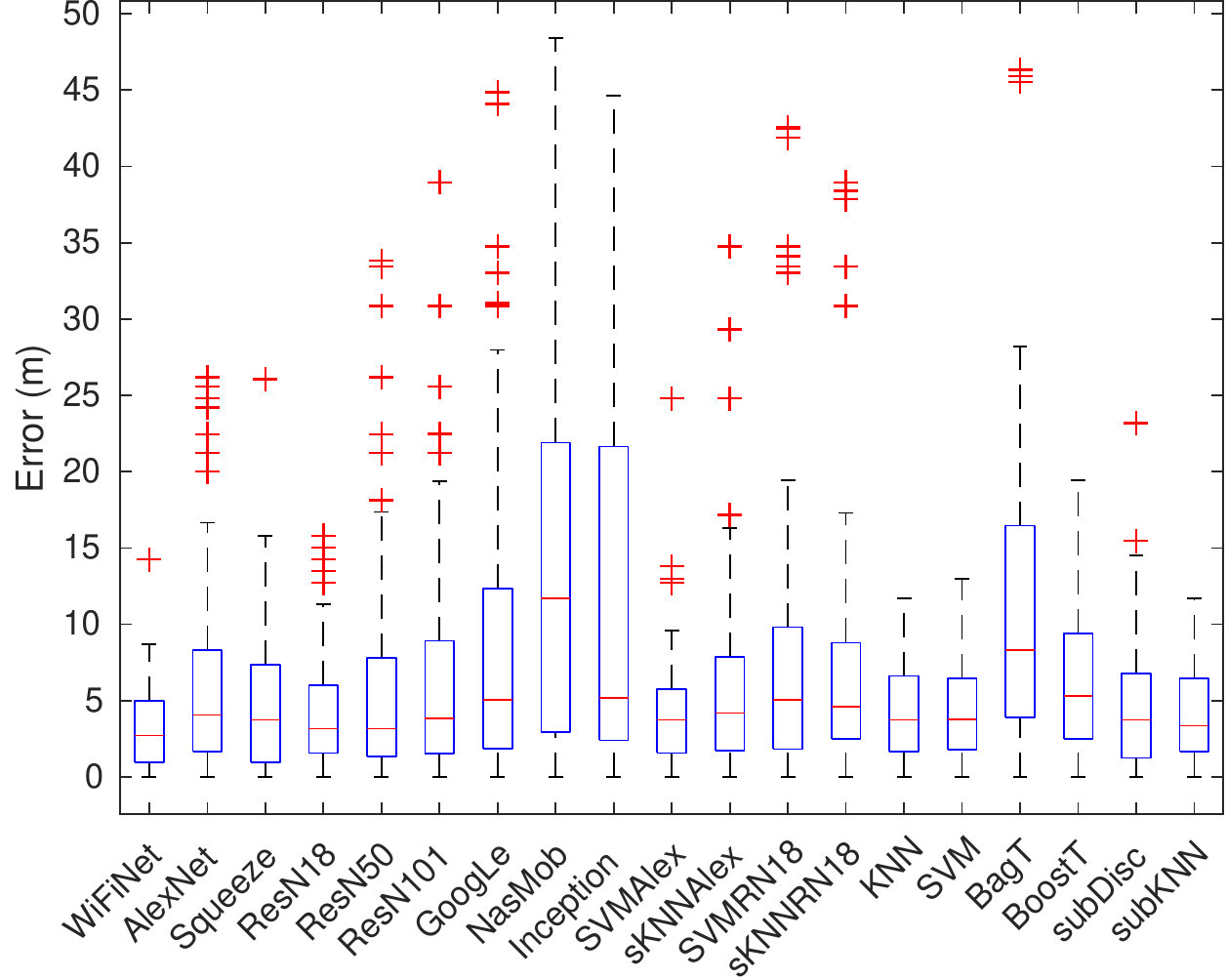}
			\label{subfig:trajBoxPlot}
		}	
		\subfigure[RMSE]{
			\includegraphics[width=0.47\linewidth] {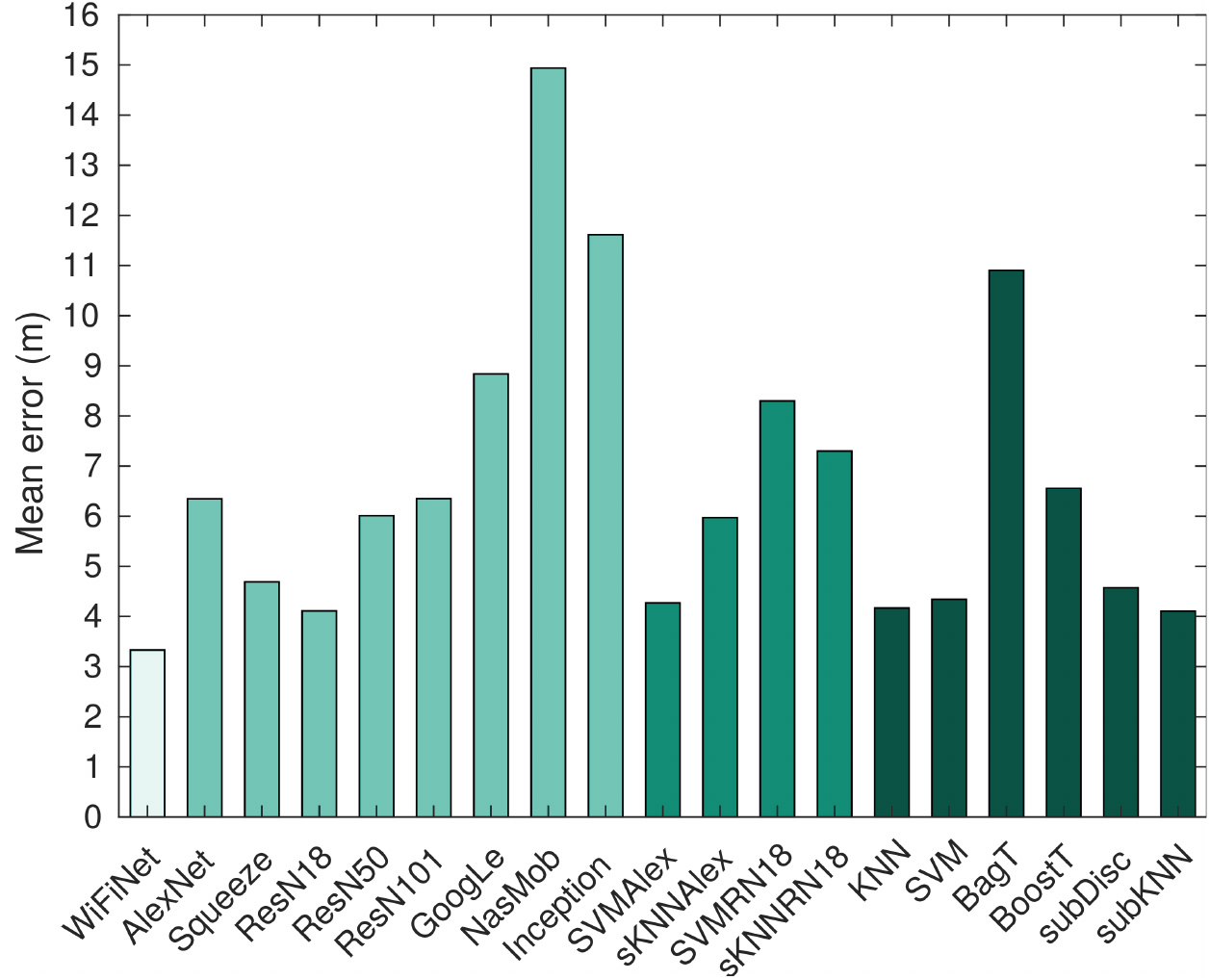}
			\label{subfig:trajMeanError}
		}	
  }
  \caption{Box plot and RMSE using the measurements collected while walking around.}
  \label{fig:trajResults}
\end{figure}

Figure \ref{fig:trajResults} shows the box plot (Figure \ref{subfig:trajBoxPlot}) and RMSE (Figure \ref{subfig:trajMeanError}) using the algorithms summarized in table \ref{tab:algorithms}. Again, four different colours are used to help with the differentiation of the four different approaches. Table \ref{tab:trajResults} shows a summary of the best results using each approach. 

As can be seen, WiFiNet is again the algorithm performing the localisation with the lowest RMSE (3.3\,m). However, the differences with the rest of the algorithms are higher in this experiment. Using ResNet18, the second best algorithm so far, the RMSE is 24\% higher and using SVM a 32\% higher than using WiFiNet. At the same time, the 75th percentile using WiFiNet (5.0\,m) is also lower than using the rest of the algorithms which means that the highest localisation errors are also lower using WiFiNet.

As a conclusion, we can say that using WiFiNet reduces the RMSE when compared with other state-of-the-art methods when locating a device in motion. We can also conclude that even though TL (especially using ResNet18) provided almost the best results when locating the device at static positions of the environment (both known and unknown), they do not seem to adapt well to motion conditions.

\begin{table}[!ht]
%   \renewcommand{\arraystretch}{1.5}
  \caption{Summary of the results locating a device in motion.}
  \label{tab:trajResults}
  \begin{center}
    \begin{tabular}{ccc} 
      \specialrule{.2em}{.1em}{.1em} 
	  Algorithm & Mean Error & $75^{th}$ percentile\\
       \specialrule{.2em}{.1em}{.1em} 
	 WiFiNet  & 3.3 m & 5.0 m \\
       
	 ResNet18 (TL)  & 4.1 m & 6.0 m \\
       
	 SVMAlex (FE)  & 4.3 m & 5.7 m \\
       
	 SVM  & 4.3 m & 6.4 m\\
      \specialrule{.2em}{.1em}{.1em} 
    \end{tabular}
  \end{center}
\end{table}

\subsection{Processing time: Localisation in real time}

Another important characteristic that was not evaluated so far is the processing time. Here, we need to differentiate between the time required to train the system and the time spent locating one sample. Of course, the training time is important, but is not a requirement to achieve real-time localisation as the training is performed off-line and prior to the use of the localisation system.

On the other hand, the time required to obtain one location estimation is critical, as the longer it takes the further the difference between two consecutive locations of the user and, as a consequence, the harder to track its position in the environment. Thus, the processing time must be lower than the WiFi acquisition time (which is usually up to 4\,Hz - 250\,ms) to achieve real-time localisation.

\begin{figure}[!hpt]
  \centering
 
	\includegraphics[width=0.5\linewidth] {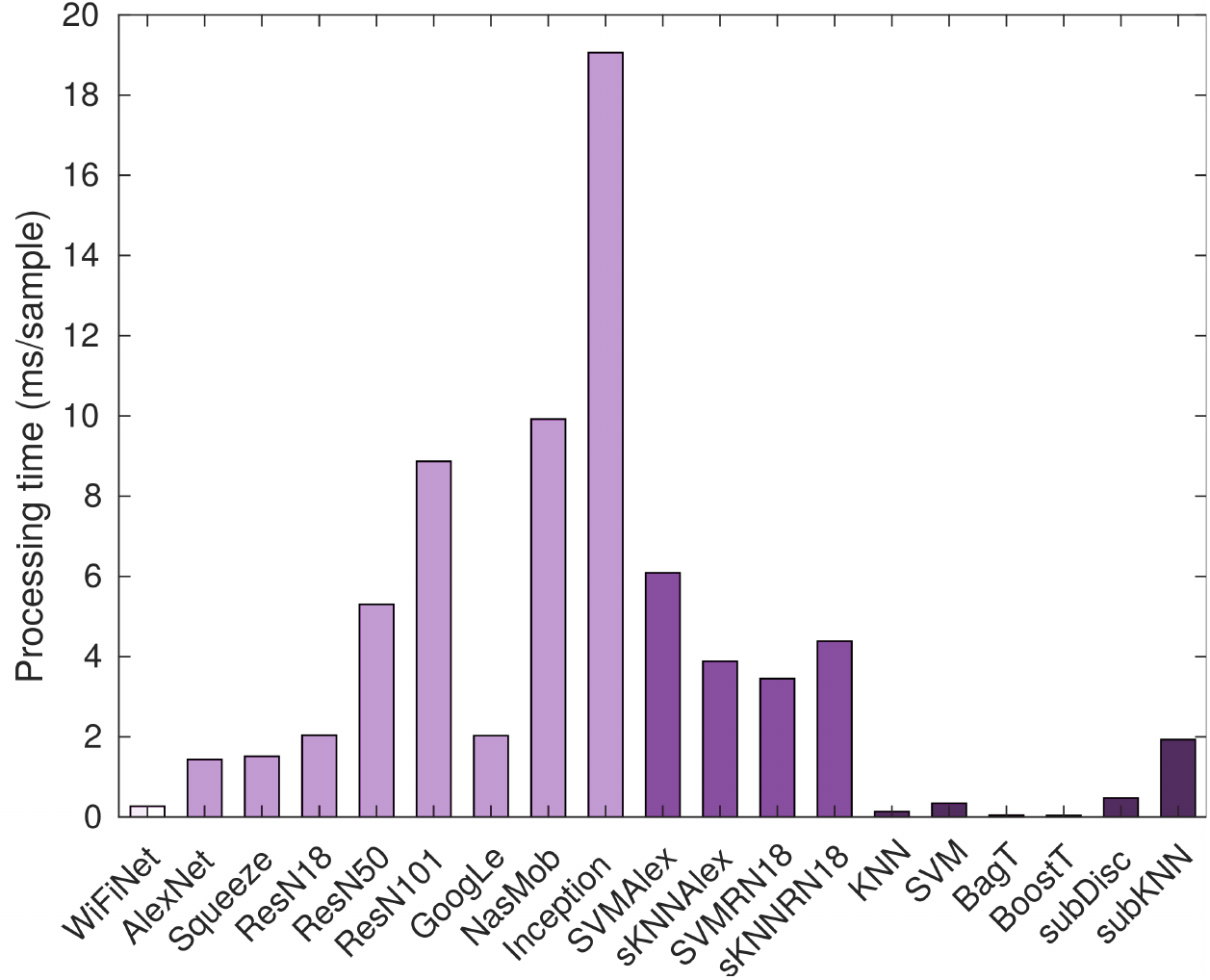}
	\label{subfig:cdf}

  \caption{Processing time per sample during localisation.}
  \label{fig:time}
\end{figure}

Figure \ref{fig:time} shows the time required to classify one sample using the different algorithms. Real-time localisation is achieved by all the algorithms as the processing times for one sample are under 20\,ms for all of them in a medium-size environment with 30 positions and 113 APs.  

However, the processing time is highly dependent on the number of positions and the number of APs. As a consequence, the algorithms might not be scalable to bigger environments
even though all the tested algorithms are able to work under real-time conditions in this scenario.

On the one hand, a higher number of APs will lead to a higher number of inputs in the classic classifiers and bigger input images for the CNNs. If we evaluate the system increasing the number of APs to 50172 (to match the pre-trained CNNs that use 224x224 images), WiFiNet needs 0,46\,ms (twice the time required using 113 APs) while SVM needs 47,27\,ms per sample (163 times the time required using 113 APs). 

On the other hand, increasing the number of positions will lead to a higher number of outputs in the classifiers. This barely affects the CNNs' processing time, as the only difference will be the connections on the last fc+softmax layers. However, for the classic classifiers an increase in the number of positions will lead to a high increase in the processing time. For an environment with 94 positions WiFiNet needs 0.27\,ms per sample (the same as in the 30 positions environment) while SVM needs 3.79\,ms per sample (13 times the time required to locate 1 sample in the 30 positions environment). 

Thus, WiFiNet is not only the localisation algorithm obtaining the lowest mean localisation error but the time it requires per location request is lower, leading us to think that it will be able to achieve real-time localisation in bigger environments than state-of-the-art methods such as SVM.

\section{Conclusions}\label{section:conclusions}

In this article we have presented a new method to estimate the location of a device using the WiFi RSS in indoor environments using CNNs. Three different approaches were evaluated: A new architecture, called WiFiNet,  designed specifically to solve the problem and the most popular pre-trained networks using both transfer learning and feature extraction. In addition, the top performing classic algorithms for WiFi-based indoor localisation were tested to be used as a baseline.

Three experiments under different conditions were carried out to analyse the performance of the different methods: Using data collected in the same positions as in the training dataset to test their accuracy, using data collected in positions non existent in the training dataset to evaluate their generalisation ability and using data collected while walking around to assess their performance under realistic conditions.

After analysing the results, we have reached the following conclusions:

\begin{itemize}

\item WiFiNet and ResNet18 (TL) are the best methods at learning how to identify static positions (both known and unknown).
\item ResNet18 (TL) and SVM show a good generalisation ability when locating the device at static positions. However, they do not seem to be as good when locating the device under realistic conditions (in motion). 
\item WiFiNet is the best generalising and adapting to realistic conditions, achieving a RMSE of 3.3 m using measurements collected while walking around the environment. 
\item Even though TL (especially using ResNet18) provided almost the best results when locating the device at static positions of the environment (both known and unknown), TL does not seem to adapt well to motion conditions.
\item Feature extraction does not seem to be a good approach especially regarding its low generalisation ability.

\end{itemize}

On the light of the results, we can conclude that using WiFiNet to perform WiFi-based indoor localisation arises as a good approach, reducing the RMSE when compared with other state-of-the-art methods, especially when locating a device under realistic conditions (3.3\,m WiFiNet vs 4.4\,m SVM). Moreover, WiFiNet is highly scalable, leading us to think that it will be able to achieve real-time localisation in bigger environments than other state-of-the-art methods such as SVM (0.46\,ms per sample WiFiNet vs 47.27\,ms per sample SVM using 50172 APs).

In the future, we plan on taking advantage of the great scalability of WiFiNet to improve localisation in bigger environments utilising the data collected in projects such as Radiocells.org \citep{references:radiocells} and obtaining ``virtual'' RSS samples for intermediate positions. To do so, we will use the CSE algorithm proposed in \citep{references:sensors17}. This way, the resolution of the WiFiNet localisation system will be improved without the need of collecting measurements on additional positions. With this approach we expect to reduce the mean localisation error without increasing the processing time which was the main problem of the original CSE algorithm and other state-of-the-art methods.

% \section{Appendices}
% 
% If there is more than one appendix, they should be identified as A, B, etc. Formulae and equations in
% appendices should be given separate numbering: Eq. (A.1), Eq. (A.2), etc.; in a subsequent appendix,
% Eq. (B.1) and so on. Similarly for tables and figures: Table A.1; Fig. A.1, etc.

\section*{Acknowledgement}

This work has received funding from the Spanish Ministry of Economy under project DPI2017-90035-R and from the European Union's Horizon 2020 research and innovation programme under the Marie Sk\l{}odowska-Curie grant agreement No 754382. The Titan Xp used for this research was donated by the NVIDIA Corporation through the GPU Grant Program.

\section*{References}

\bibliography{2019_ESWA_NHernandez}

\end{document}